%% file: main.tex
\documentclass{article}

\usepackage{microtype}
\usepackage{graphicx}
\usepackage{subcaption}
\usepackage{booktabs}
\usepackage{hyperref}
\usepackage{amsmath}
\usepackage{amssymb}
\usepackage{tikz}
\usetikzlibrary{
  positioning,
  arrows.meta,
  calc,
  fit,
  backgrounds,
  decorations.pathreplacing,
  shapes.geometric,
}
\usepackage{mathtools}
\usepackage{amsthm}
\usepackage[capitalize,noabbrev]{cleveref}
\usepackage{placeins} 
\usepackage{afterpage}

\usepackage[accepted]{icml2026}
\usepackage[bottom]{footmisc}

\icmltitlerunning{PRISM: Autoregressive Inverse Design of Multilayer Thin Films}

\begin{document}

\twocolumn[
  \icmltitle{PRISM: Position-encoded Regressive Inverse Spectral Model\\for Multilayer Thin-Film Design}

  \icmlsetsymbol{equal}{*}

  \begin{icmlauthorlist}
    \icmlauthor{Runtian Wang}{equal,ind}
    \icmlauthor{Renhao Xue}{equal,ind,amz}
    \icmlauthor{Baige Chen}{ind}
    \icmlauthor{Hao Wu}{ind}
  \end{icmlauthorlist}

  \icmlaffiliation{ind}{Independent Researcher}
  \icmlaffiliation{amz}{Work does not relate to position at Amazon}
  \icmlcorrespondingauthor{Runtian Wang}{w.henry4@live.com}
  \icmlcorrespondingauthor{Renhao Xue}{lukasxue@amazon.com}
  \vskip 0.3in

  \begin{abstract}
  \input{sections/abstract}
  \end{abstract}
]

\printAffiliationsAndNotice{\icmlEqualContribution}

\input{sections/introduction}
\input{sections/related_work}
\input{sections/method}
\input{sections/experiments}
\input{sections/results}
\input{sections/analysis}

\input{sections/conclusion}

\bibliography{references}
\bibliographystyle{icml2026}

\end{document}

%% file: sections/abstract.tex
The inverse problem of multilayer thin-film optical coatings design represents a complex combinatorial-continuous optimization challenge. We present \textbf{PRISM} (Position-encoded Regressive Inverse Spectral Model), a unified decoder-only autoregressive transformer that streamlines this process by jointly predicting discrete material selection and continuous thickness regression within a single backbone. PRISM introduces two primary architectural innovations: (1) \textit{spectrum prefix conditioning}, which utilizes standard prefix tokens for in-context target injection, and (2) \textit{cumulative-depth Rotary Position Embeddings}, which encode continuous thickness directly into the positional representation to preserve the physical spatial relationships of the stack. Our benchmarks demonstrate that a PRISM-13M model reduces MAE by over 50\% compared to other transformer baselines while utilizing only one-fifth of the parameters. Furthermore, a 44M-parameter variant achieves state-of-the-art performance (MAE = 0.010) on our in-distribution validation benchmark and operates significantly faster than simulated annealing, offering a highly efficient alternative to classical optimization methods.

%% file: sections/introduction.tex
\section{Introduction}
\label{sec:introduction}

Multilayer thin-film optical coatings are ubiquitous in modern photonics. Anti-reflection coatings, bandpass filters, dichroic mirrors, and neutral-density filters all rely on precisely engineered stacks of dielectric and metallic layers whose interference effects shape the spectral response~\citep{macleod2010thin}. The \emph{forward} problem of computing the reflectance and transmittance spectrum of a given stack is solved analytically by the Transfer Matrix Method (TMM)~\citep{born2019principles}. The \emph{inverse} problem on the other hand, is far harder. It is a mixed discrete-continuous optimization over material choices (combinatorial) and layer thicknesses (continuous), with a highly non-convex landscape and many degenerate solutions.

Traditional approaches to inverse thin-film design rely on iterative numerical optimization. Simulated annealing~\citep{kirkpatrick1983optimization}, genetic algorithms~\citep{martin1995synthesis}, and needle optimization~\citep{tikhonravov1996application} search the design space directly, evaluating thousands of candidate structures via TMM at test time. While these methods can achieve high spectral fidelity, they are computationally expensive, requiring minutes to hours per design, making them impractical for real-time or high-throughput applications.

Recent neural approaches have sought to amortize the cost of inverse design with varying success. Tandem networks \citep{liu2018training} utilize joint MLP-surrogate training but are limited by fixed-length representations, while conditional generative models \citep{so2019designing} often struggle with mode coverage. 

Transformer-based sequence modeling has emerged as a promising alternative, yet current frameworks, OptoGPT \citep{ma2024optogpt} and OptoFormer \citep{wu2026optoformer}, face important efficiency and precision bottlenecks. OptoGPT is a decoder-only autoregressive transformer that serializes each layer into a joint material-thickness token, which requires discretizing thickness and creates a large joint vocabulary. OptoFormer instead uses an encoder with a dual-decoder architecture to separately predict material and thickness sequences, reducing the joint-vocabulary issue but adding architectural complexity. PRISM addresses these limitations by combining factored material prediction and continuous thickness regression within a single decoder-only backbone. We explore these limitations in detail in \Cref{sec:related}.

We propose \textbf{PRISM} (Position-encoded Regressive Inverse Spectral Model), an autoregressive transformer that addresses these limitations. PRISM generates thin-film designs layer by layer using causal self-attention, with two key innovations:

\begin{itemize}
    \item \textbf{Spectrum prefix conditioning} (\Cref{sec:prefix}): The target spectrum is projected through a linear layer into a single token prepended to the decoder sequence. The entire model uses causal self-attention only, simplifying the architecture while keeping the conditioning always visible in the attention window.
    \item \textbf{Cumulative-depth RoPE} (\Cref{sec:rope}): Rotary Position Embeddings~\citep{su2024roformer} use the cumulative physical depth of the film stack (in nm) rather than sequential token index. This gives the attention mechanism a physically meaningful distance metric that directly reflects the geometric depth in the stack, providing an elegant solution to the problem of continuous thickness input for the transformer architecture.
\end{itemize}

These innovations are realized in a unified decoder backbone with a per-material thickness regression head, enabling joint beam search over (material, thickness) pairs at inference.

We evaluate PRISM across a broad design space of 17 dielectric and metallic materials. Our results demonstrate that PRISM achieves state-of-the-art accuracy while maintaining high parameter efficiency. Our models substantially outperform both neural baselines and simulated annealing on in distribution benchmarks while staying competitive against traditional methods on practical targets with much faster inference.

%% file: sections/related_work.tex
\section{Related Work}
\label{sec:related}

\paragraph{Classical inverse thin-film design.}
The inverse design of multilayer optical coatings has a long history in optical engineering. Needle optimization~\citep{tikhonravov1996application} iteratively inserts thin layers at positions that maximally reduce the merit function, then locally optimizes thicknesses. Genetic algorithms~\citep{martin1995synthesis} and simulated annealing~\citep{kirkpatrick1983optimization} perform global stochastic search over the joint discrete-continuous design space. Gradient-based methods using differentiable TMM implementations~\citep{Luce2022TMMFast} enable L-BFGS optimization of thicknesses for fixed material sequences. All these methods optimize directly against each target spectrum at test time, achieving high fidelity at the cost of minutes to hours per design.

\paragraph{Neural inverse design in nanophotonics.}
Neural networks have been applied to inverse design across nanophotonics, including metasurfaces~\citep{jiang2019free}, plasmonic nanostructures~\citep{malkiel2018plasmonic}, and thin films. Tandem networks~\citep{liu2018training} address the one-to-many nature of inverse problems by jointly training an inverse network with a forward surrogate that enforces spectrum consistency. However, they use fixed-length representations (e.g., padded to a maximum layer count) and cannot naturally handle variable-length designs. Conditional generative models, including conditional GANs~\citep{so2019designing} and CVAEs~\citep{sohn2015learning}, learn distributions over designs conditioned on target spectra but often struggle with mode coverage and require multiple samples for good results.

\paragraph{Autoregressive models for inverse design.}
OptoGPT~\citep{ma2024optogpt} pioneered the use of autoregressive transformers~\citep{vaswani2017attention} for thin-film inverse design, treating the problem as sequence generation. It uses a decoder-only autoregressive architecture with spectrum conditioning and a joint material-thickness vocabulary, where each token encodes both a material and a discretized thickness. While effective, this approach has two drawbacks: (i)~the vocabulary scales as $|\text{materials}| \times |\text{thickness bins}|$, reaching 904 tokens for 18 materials and 50 bins; and (ii)~thickness precision is limited to the bin width (10\,nm).

\paragraph{Concurrent work.}
OptoFormer~\citep{wu2026optoformer}, developed concurrently and independently, shares the motivation of moving beyond joint material-thickness vocabularies. It addresses the vocabulary explosion by factoring generation into separate material and thickness streams via a dual-decoder architecture: a spectrum encoder produces a latent representation consumed by two independent decoders. This separation avoids the combinatorial vocabulary but introduces substantial architectural complexity through an encoder and two decoders. PRISM arrives at a different solution, achieving both factored prediction and continuous thickness regression within a single decoder backbone. As neither code nor pretrained models for OptoFormer are publicly available, we were unable to include it as a baseline; we compare against it qualitatively on architectural design choices.

\paragraph{Rotary Position Embeddings.}
RoPE~\citep{su2024roformer} encodes position information by rotating query and key vectors in the complex plane, with rotation angles proportional to position. Originally designed for sequential token positions, RoPE has been extended to longer contexts via Position Interpolation~\citep{chen2023extending}, NTK-aware scaling~\citep{bloc2023ntk}, and YaRN~\citep{peng2024yarn}. Our work repurposes RoPE for a non-sequential domain: we use cumulative physical depth (in nm) as the position, giving the attention mechanism a distance metric grounded in the physics of thin-film interference.

%% file: sections/method.tex
\section{Method}
\label{sec:method}

\subsection{Problem Formulation}
\label{sec:problem}

A multilayer thin-film stack consists of $L$ layers, each specified by a discrete material $m_\ell$ and positive continuous thickness $d_\ell$ (in nm), deposited on a glass substrate. The forward model $f$ computes the optical spectrum consisting of reflectance and transmittance via the Transfer Matrix Method (TMM)~\citep{born2019principles}:
\begin{equation}
    \mathbf{s} = f\bigl((m_1, d_1), \ldots, (m_L, d_L)\bigr) \in [0,1]^{142},
\end{equation}
where $\mathbf{s}$ is the list of concatenated reflectance and transmittance values sampled at set intervals across the EM spectrum. The inverse problem seeks a design $(m_1, d_1), \ldots, (m_L, d_L)$ whose spectrum $f(\cdot)$ matches a given target $\mathbf{s}^*$.

We frame this as auto-regressive sequence generation. Given $\mathbf{s}^*$, the model generates layers left to right:
\begin{equation}
    p\bigl((m_\ell, d_\ell) \mid \mathbf{s}^*, (m_1, d_1), \ldots, (m_{\ell-1}, d_{\ell-1})\bigr),
\end{equation}
terminating when it emits an \texttt{EOS} token. Material selection is a categorical distribution over $17$ materials; thickness is a continuous regression target.

\subsection{Architecture Overview}

\begin{figure*}[t]
\centering
\begin{subfigure}[b]{0.58\textwidth}
    \centering
    \input{figures/architecture_diagram.tex}
    \caption{PRISM architecture. (A)~The target spectrum is projected into a single prefix token via a linear layer. (B)~Material tokens are embedded via learned embeddings and the spectrum embedding is appended as the first token. (C)~Cumulative-depth RoPE encodes positions as the running sum of layer thicknesses in nm. The shared transformer backbone feeds two output heads: a linear material head producing next-material logits, and a thickness regression head predicting continuous thicknesses.}
    \label{fig:architecture}
\end{subfigure}
\hfill
\begin{subfigure}[b]{0.40\textwidth}
    \includegraphics[width=\linewidth]{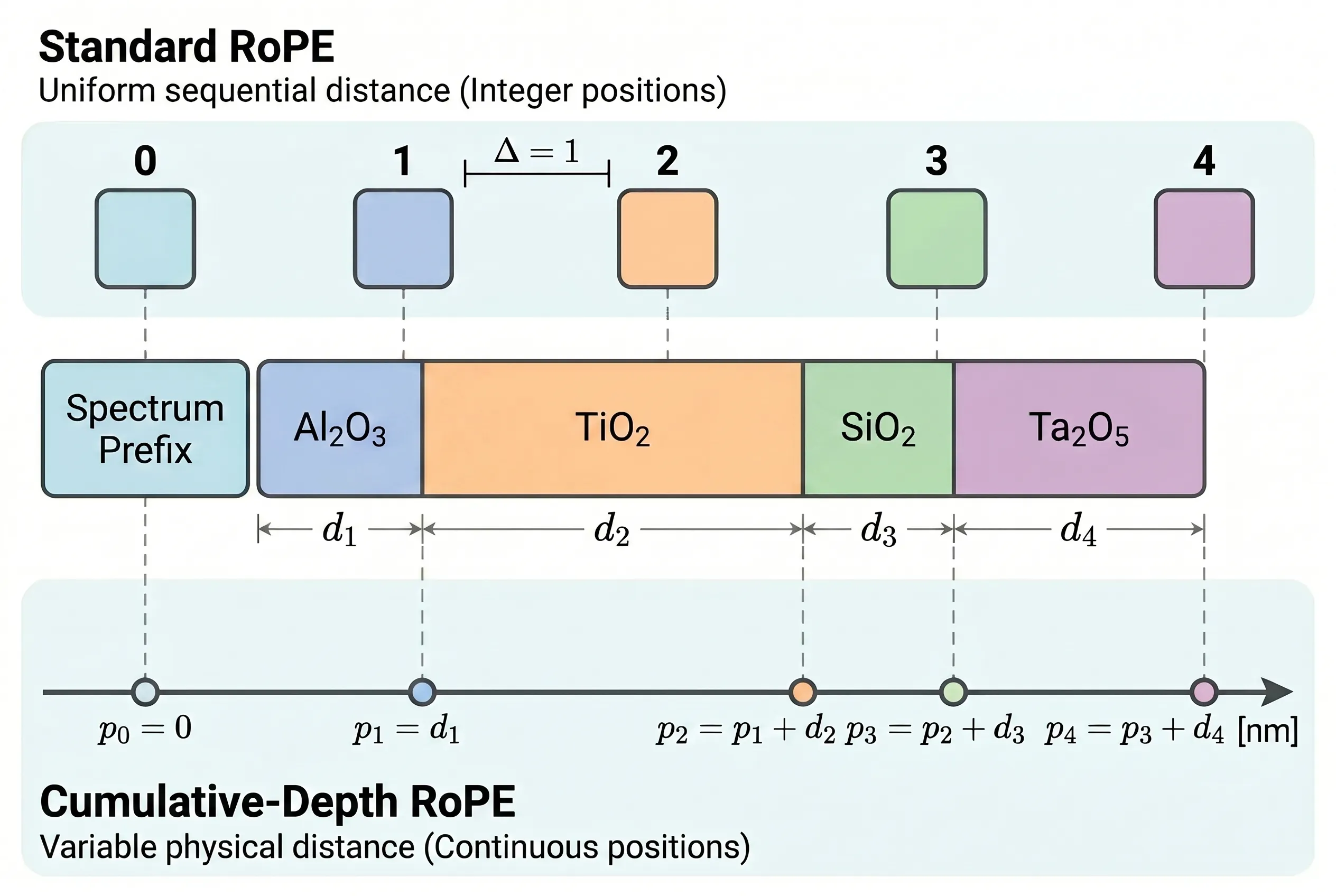}
    \caption{Cumulative-depth RoPE. A multilayer stack with per-layer thicknesses $d_1, \ldots, d_L$ and cumulative depths $p_0, \ldots, p_L$. Standard RoPE assigns integer positions $0, 1, \ldots, L$ to the embedding sequence. Cumulative-depth RoPE assigns each token a continuous position equal to the running sum of thicknesses, aligning the attention geometry with physical separation in nm.}
    \label{fig:cumdepth_rope}
\end{subfigure}
\caption{PRISM architecture (left) and cumulative-depth RoPE positional encoding (right).}
\label{fig:method_overview}
\end{figure*}

PRISM is a decoder-only transformer with a single shared backbone feeding two output heads (\Cref{fig:architecture}). The input sequence is:
\begin{equation}
    [\,\underbrace{\texttt{SPEC}}_{\text{prefix}},\; \underbrace{(m_1, d_1),\; \ldots,\; (m_{L}, d_{L}),\; \texttt{EOS}}_{\text{layer tokens}}\,],
\end{equation}
where \texttt{SPEC} is a learned projection of the target spectrum and each layer token is embedded from its material ID only (thickness is encoded via positional embeddings).

\subsection{Spectrum Prefix Conditioning}
\label{sec:prefix}

The target spectrum $\mathbf{s}^*$ is projected into a single $d_\text{model}$-dimensional token via a linear layer:
\begin{equation}
    \mathbf{h}_{\text{spec}} = W_{\text{spec}}\, \mathbf{s}^* + \mathbf{b}_{\text{spec}}.
\end{equation}
This token is prepended to the decoder sequence. Under causal masking, the spectrum prefix attends only to itself, while all subsequent tokens attend to the prefix and to all preceding tokens. This replaces the encoder and cross-attention of standard encoder-decoder designs with a simpler architecture as the conditioning is always within the causal attention window.

\subsection{Cumulative-Depth RoPE}
\label{sec:rope}

Standard RoPE~\citep{su2024roformer} assigns integer positions $0, 1, 2, \ldots$ to tokens. In thin-film design, the physically relevant quantity is not the layer index but the cumulative geometric depth. Two layers separated by a thick intervening layer interact differently (via interference) than two layers in close proximity, even if they are adjacent in the sequence.

We define the position of each token as the cumulative thickness up to that layer:
\begin{equation}
    p_0 = 0, \quad p_\ell = \sum_{j=1}^{\ell} d_j, \quad \ell = 1, \ldots, L.
\end{equation}
The spectrum prefix sits at position $p_0 = 0$. These continuous positions are used directly in place of integer indices in the standard RoPE formulation~\citep{su2024roformer}, applied to both queries and keys in every attention layer. \Cref{fig:cumdepth_rope} contrasts this scheme with standard integer-indexed RoPE.

This encoding has two advantages: (1)~the attention dot product between two tokens depends on their \emph{physical separation} in nm, not their sequential distance, aligning the inductive bias with thin-film interference physics; and (2)~the model can naturally handle variable total depths without retraining, since positions are continuous rather than bounded integers.

\subsection{Dual Output Heads}
\label{sec:thkhead}

The shared transformer backbone produces hidden states $\mathbf{h}_\ell \in \mathbb{R}^{d_\text{model}}$ at each position. The vocabulary $V$ comprises the 17 materials plus the special tokens \texttt{PAD} and \texttt{EOS}. Two heads operate on the hidden states:

\paragraph{Material head.} A linear projection produces logits over the vocabulary:
\begin{equation}
    \mathbf{z}_\ell^{\text{mat}} = W_{\text{mat}}\, \mathbf{h}_\ell \in \mathbb{R}^{|V|}.
\end{equation}

\paragraph{Per-material thickness head.} A multi-layer MLP predicts one thickness for every entry of the vocabulary simultaneously:
\begin{equation}
    \hat{\mathbf{d}}_\ell = \text{THK\_MIN} \cdot \exp\bigl(\text{softplus}\bigl(\text{MLP}(\mathbf{h}_\ell)\bigr)\bigr) \in \mathbb{R}_{>0}^{|V|}
\end{equation}
The MLP consists of two hidden layers with GELU activations and dropout. The softplus activation ensures positivity in log-space; the exponential maps back to nm. The output dimension matches the material head for indexing convenience; only entries corresponding to physical materials are used at decode time. When material $m$ is selected at position $\ell$, the prediction $\hat{d}_\ell^{(m)}$ is used.

This design is the key enabler for \textbf{beam search}. The augmented regression head  allows the model to predict thickness values conditioned on the material token chosen for the layer. This avoids the two-stage decode required when thickness depends on the chosen material.

\subsection{Training}
\label{sec:training}

\paragraph{Material loss.} We use label-smoothed KL divergence~\citep{szegedy2016rethinking} with smoothing $\epsilon = 0.1$:
\begin{equation}
    \mathcal{L}_{\text{mat}} = \text{KL}\bigl(\tilde{p} \,\|\, \hat{p}\bigr), \quad \tilde{p}_i = \begin{cases} 1 - \epsilon & i = y \\ \epsilon / (V - 2) & i \neq y, i \neq \texttt{PAD} \end{cases},
\end{equation}
where $\hat{p}$ is the softmax of material logits and $y$ is the ground-truth material.

\paragraph{Thickness loss.} We compute MSE in log-space, masked to non-padding positions and gathered at the ground-truth material index:
\begin{equation}
    \mathcal{L}_{\text{thk}} = \sum_{\ell : y_\ell \neq \texttt{PAD}} \bigl(\tilde{d}_\ell^{(y_\ell)} - \log(d_\ell / \text{THK\_MIN})\bigr)^2,
\end{equation}
where $\tilde{d}_\ell^{(y_\ell)}$ is the log-space prediction at the ground-truth material index. Log-space training prevents large thicknesses from dominating the loss.

\paragraph{Total loss.} Both $\mathcal{L}_{\text{mat}}$ and $\mathcal{L}_{\text{thk}}$ are sums over non-padding positions; the combined loss is normalised once by the number of non-padding tokens:
\begin{equation}
    \mathcal{L} = \frac{\mathcal{L}_{\text{mat}} + \alpha \cdot \mathcal{L}_{\text{thk}}}{N},
\end{equation}
where $\alpha$ is the thickness loss weight and $N$ is the number of non-padding tokens.

\subsection{Decoding and Re-ranking}
\label{sec:decoding}

At inference, PRISM supports greedy decoding and beam search, as enabled by the per-material thickness head described in \Cref{sec:thkhead}.

Crucially, during beam search, decoded designs are \textbf{re-simulated via TMM} to obtain physically accurate spectra. This re-simulation enables a powerful \textbf{TMM-reranking} strategy: rather than relying solely on model log-probability to select among beam candidates, we rank the $K$ candidates by their true spectral error against the target. Since TMM evaluation is cheap, re-ranking adds negligible cost while selecting for physical fidelity rather than model confidence. This allows the model to benefit from diversity in its beam since a high-quality design need only appear among the $K$ candidates to be selected, even if the model assigns it lower probability than a spectrally inferior alternative.

%% file: figures/architecture_diagram.tex

\definecolor{specblue}{HTML}{4A90D9}
\definecolor{specbluelt}{HTML}{D6E6F5}
\definecolor{matgreen}{HTML}{5BA05B}
\definecolor{matgreenlt}{HTML}{D4EDCF}
\definecolor{thkorange}{HTML}{E8913A}
\definecolor{thkorangelt}{HTML}{FDEBD0}
\definecolor{backbone}{HTML}{5B6B8A}
\definecolor{backbonelt}{HTML}{D8DDE6}
\definecolor{ropepurple}{HTML}{8E6BAD}
\definecolor{ropepurplelt}{HTML}{E8DAEF}
\definecolor{headred}{HTML}{C0504D}
\definecolor{headredlt}{HTML}{F2DCDB}
\definecolor{headamber}{HTML}{D4A437}
\definecolor{headamberlt}{HTML}{FDF2D0}
\definecolor{lightgray}{HTML}{F4F4F4}
\definecolor{midgray}{HTML}{AAAAAA}

\resizebox{\linewidth}{!}{%
\begin{tikzpicture}[
  >=Stealth,
  font=\footnotesize,
  node distance=0.4cm,
  every node/.style={align=center},
  block/.style={
    draw, rounded corners=5pt, minimum height=0.6cm,
    minimum width=1.8cm, line width=0.6pt,
  },
  smallblock/.style={
    draw, rounded corners=4pt, minimum height=0.5cm,
    minimum width=1.0cm, line width=0.5pt, font=\scriptsize,
  },
  arrow/.style={->, line width=0.6pt, color=black!70},
  darrow/.style={->, line width=0.8pt, color=black!80},
  label/.style={font=\scriptsize\itshape, text=black!60},
  seclabel/.style={font=\small\bfseries, text=black!80},
]


\node[smallblock, fill=specbluelt, draw=specblue, minimum width=2.4cm]
  (spectrum) {Target Spectrum};

\node[label, below=0.03cm of spectrum]
  (speclabel) {\scriptsize Reflectance \& Transmittance};

\node[block, fill=specblue!20, draw=specblue, above=0.45cm of spectrum, minimum width=3.7cm]
  (specproj) {Linear Projection};

\draw[arrow] (spectrum) -- (specproj);

\node[smallblock, fill=specbluelt, draw=specblue, above=0.4cm of specproj]
  (spectok) {$\mathbf{h}_0$};

\draw[arrow] (specproj) -- (spectok);

\node[below=0.25cm of speclabel] (secA) {};


\node[right=1.2cm of spectrum] (matinput_anchor) {};

\foreach \i/\lab in {1/m_1, 2/m_2, 3/m_3, 4/\cdots, 5/m_t} {
  \pgfmathsetmacro{\xoff}{(\i-1)*0.75}
  \node[smallblock, fill=matgreenlt, draw=matgreen,
        minimum width=0.6cm, minimum height=0.45cm]
    (mat\i) at ([xshift=\xoff cm]matinput_anchor) {$\lab$};
}

\node[label, below=0.03cm of mat3] (matlabel) {\scriptsize Material IDs};

\node[block, fill=matgreen!15, draw=matgreen,
      above=0.45cm of mat3, minimum width=3.7cm]
  (tokemb) {Material Embedding};

\foreach \i in {1,...,5} {
  \draw[arrow] (mat\i) -- (mat\i |- tokemb.south);
}

\node[smallblock, fill=matgreenlt, draw=matgreen,
      above=0.4cm of tokemb, minimum width=3.7cm]
  (matembout) {$\mathbf{h}_1, \ldots, \mathbf{h}_t$};

\draw[arrow] (tokemb) -- (matembout);


\node[right=0.6cm of mat5] (rope_anchor) {};

\foreach \i/\lab in {1/0, 2/d_1, 3/d_2, 4/\cdots, 5/d_t} {
  \pgfmathsetmacro{\xoff}{(\i-1)*0.75}
  \node[smallblock, fill=thkorangelt, draw=thkorange,
        minimum width=0.6cm, minimum height=0.45cm]
    (thk\i) at ([xshift=\xoff cm]rope_anchor) {$\lab$};
}

\node[label, below=0.03cm of thk3] (thklabel) {\scriptsize Thickness (nm)};

\node[block, fill=ropepurplelt, draw=ropepurple,
      above=0.45cm of thk3, minimum width=3.7cm]
  (cumsum) {Cumulative Sum};

\foreach \i in {1,...,5} {
  \draw[arrow] (thk\i) -- (thk\i |- cumsum.south);
}

\node[smallblock, fill=ropepurple!15, draw=ropepurple,
      above=0.4cm of cumsum, minimum width=3.7cm]
  (rope) {RoPE};

\draw[arrow] (cumsum) -- (rope);


\node[draw=midgray, fill=lightgray, circle, minimum size=0.55cm,
      line width=0.5pt, font=\footnotesize, inner sep=0pt]
  (concat) at ([yshift=0.7cm]matembout.north -| mat3)
  {$\oplus$};

\draw[arrow] (spectok.north) |- (concat);
\draw[arrow] (matembout.north) -- (concat.south);

\node[draw=backbone, rounded corners=6pt, line width=0.8pt,
      fill=backbonelt!50,
      minimum width=4.0cm, minimum height=2.3cm,
      above=0.3cm of concat]
  (backbone_box) {};

\node[font=\small\bfseries, text=backbone, anchor=north]
  at ([yshift=-3pt]backbone_box.north)
  {Transformer $\times N$};

\node[block, fill=ropepurplelt, draw=ropepurple,
      minimum width=3.7cm, minimum height=0.6cm]
  (selfattn) at ($(backbone_box.south) + (0, 0.5cm)$)
  {Multi-Head Self-Attention};

\node[block, fill=backbone!8, draw=backbone,
      minimum width=3.7cm, minimum height=0.6cm,
      above=0.25cm of selfattn]
  (ffn) {FFN};


\draw[arrow] (selfattn) -- (ffn);

\draw[darrow] (concat) -- (concat |- backbone_box.south);

\draw[->, line width=0.6pt, color=ropepurple!80, dashed] (rope) |- (selfattn.east)
  node[pos=0.85, above, font=\tiny, text=ropepurple]{};


\node[block, fill=headredlt, draw=headred,
      above left=0.6cm and 0.6cm of backbone_box.north, minimum width=2.4cm]
  (mathead) {Material Head};

\node[block, fill=headamberlt, draw=headamber,
      above right=0.6cm and 0.6cm of backbone_box.north, minimum width=2.4cm]
  (thkhead) {Thickness Head};

\draw[darrow] (backbone_box.north) -- ++(0,0.12) -| (mathead.south);
\draw[darrow] (backbone_box.north) -- ++(0,0.12) -| (thkhead.south);


\begin{pgfonlayer}{background}
  \node[fit=(spectrum)(specproj)(spectok)(speclabel),
        fill=specblue!6, draw=specblue!30, dashed, rounded corners=6pt,
        inner sep=3pt] (bgA) {};
  \node[font=\scriptsize\bfseries, text=black!70, anchor=south west]
    at (bgA.south west) {(A)};

  \node[fit=(mat1)(mat5)(tokemb)(matembout)(matlabel),
        fill=matgreen!6, draw=matgreen!30, dashed, rounded corners=6pt,
        inner sep=3pt] (bgB) {};
  \node[font=\scriptsize\bfseries, text=black!70, anchor=south west]
    at (bgB.south west) {(B)};

  \node[fit=(thk1)(thk5)(cumsum)(rope)(thklabel),
        fill=ropepurple!6, draw=ropepurple!30, dashed, rounded corners=6pt,
        inner sep=3pt] (bgC) {};
  \node[font=\scriptsize\bfseries, text=black!70, anchor=south west]
    at (bgC.south west) {(C)};
\end{pgfonlayer}

\end{tikzpicture}%
}

%% file: sections/experiments.tex
\section{Experimental Setup}
\label{sec:experiments}

\subsection{Design Space}

Our design space comprises 17 materials (dielectrics: Al$_2$O$_3$, AlN, HfO$_2$, MgF$_2$, MgO, Si$_3$N$_4$, SiO$_2$, Ta$_2$O$_5$, TiO$_2$, ZnO, ZnS, ZnSe; semiconductors: Ge, Si; metals: Al, ITO, TiN) deposited on a glass substrate. Layer thicknesses range from 10 to 500\,nm in 10\,nm steps, with 1--20 layers per stack. Spectra are computed at 71 wavelengths (400--1100\,nm, 10\,nm spacing) for both reflectance and transmittance (142 values total), using TMM with incoherent substrate treatment (500\,$\mu$m glass, s-polarization, normal incidence). Refractive index data ($n + ik$) for each material is loaded from tabulated measurements and interpolated to the wavelength grid via cubic splines.

\subsection{Data Generation}

Training data is generated by uniformly sampling material sequences and thicknesses, then simulating spectra via TMM. Layer counts are sampled with probability $P(L) \propto L$ to oversample longer sequences. We generate up to 30M training samples, 100K development samples, and 10K validation samples.

\subsection{Model Configuration}

We train two models to study the effect of scale:

\textbf{PRISM-13M.} $d_\text{model} = 256$, $d_\text{ff} = 1024$, 4 attention heads, 4 transformer layers, dropout 0.1, 2-hidden-layer thickness MLP head. Trained for 30 epochs on 10M samples to enable direct comparison with OptoGPT under matched data conditions.

\textbf{PRISM-44M.} $d_\text{model} = 768$, $d_\text{ff} = 3072$, 8 attention heads, 6 transformer layers, dropout 0.1, 2-hidden-layer thickness MLP head. Trained for 60 epochs on 30M samples for maximum performance.

Both models use batch size 1024, thickness loss weight $\alpha = 1.0$, and AdamW optimizer~\citep{loshchilov2019decoupled} with cosine annealing learning rate schedule~\citep{loshchilov2017sgdr}.

\subsection{Baselines}

We compare against five baselines spanning optimization and neural approaches, all evaluated on the same validation set and practical targets with identical TMM re-simulation:

\textbf{Simulated Annealing (SA)~\citep{kirkpatrick1983optimization}.} Stochastic global optimization with 8 restarts $\times$ 5{,}000 steps each. Moves include thickness perturbation ($\sigma = 30$\,nm), material swap, layer insertion, and layer removal. Exponential temperature schedule from $T = 0.1$ to $10^{-4}$.

\textbf{Differentiable TMM (Diff-TMM)~\citep{Luce2022TMMFast}.} Gradient-based optimization via a PyTorch-differentiable TMM implementation. L-BFGS with 32 random restarts across layer counts $\{3, 5, 7, 10, 14, 18\}$, 300 iterations per restart. Thicknesses parameterized in log-space.

\textbf{OptoGPT~\citep{ma2024optogpt}.} Pretrained 63.9M-parameter autoregressive transformer with cross-attention. Joint material-thickness vocabulary with approximately 900 structure tokens. We use the published checkpoint (epoch 146) with greedy decoding.

\textbf{Tandem Network~\citep{liu2018training}.} Joint inverse-forward MLP (1.58M parameters). The inverse network predicts fixed-length (20-layer) material logits, thicknesses, and layer count; the forward network reconstructs the spectrum for consistency loss. Trained on 10M samples for 30 epochs.

\textbf{CVAE.} Conditional VAE (1.41M parameters, 64-dim latent), following the generative design approach of~\citet{so2019designing}. Encoder maps (spectrum, structure) to latent distribution; decoder generates fixed-length designs from spectrum + sampled $z$. KL weight annealed from 0 to 0.1.

\subsection{Evaluation Protocol}

\textbf{Metrics.} All metrics are computed over the full 142-dimensional spectrum vector after TMM re-simulation of predicted designs. We report Mean Absolute Error (MAE) and coefficient of determination ($R^2$) on both benchmarks. On the practical targets benchmark we additionally report a spectral earth-mover's distance (EMD), a shape-sensitive metric that measures the minimum cost to transport mass between the predicted and target spectra along the wavelength axis (computed independently on the reflectance and transmittance components and summed). EMD complements pointwise MAE by tolerating small wavelength shifts of correctly-shaped spectral features. Importantly, we never compare predicted structures to ground-truth structures directly.

\textbf{Decoding configurations.} For PRISM, we report two decoding variants: (1)~greedy, (2)~TMM-reranked-best (beam seach with spectrum error re ranking)

\textbf{Benchmarks.} We evaluate on two benchmark categories:
\begin{itemize}
    \item \textbf{Generated targets (in-distribution)}: 10{,}000 randomly generated structures with 1--20 layers, matching the training distribution.
    \item \textbf{Practical targets}: 84 practical optical filter spectra spanning a broad range of categories (narrowband, broadband, edge, notch, bandstop, dichroic, multi-bandpass, hot/cold mirrors, broadband HR mirrors, beam splitters, linear variable filters, color filters, and other specialty filters), which do not appear in the training distribution.
\end{itemize}

%% file: sections/results.tex
\section{Results}
\label{sec:results}

\subsection{In-Distribution Performance}
\label{sec:results_synthetic}

\Cref{tab:main_results} compares PRISM at both scales against all baselines on both the validation set (in-distribution) and the 84 practical targets (out-of-distribution). The practical set comprises practical filter spectra whose shapes are distributionally distinct from the training data, testing whether each method can generalize to practical optical designs.

\begin{table*}[!t]
\caption{Comparison on in-distribution validation and out-of-distribution practical targets. All spectra are TMM re-simulated. $^\dagger$Optimization methods evaluated on $n = 1{,}000$ on validation due to computational cost. Practical target evaluation uses all 84 targets. PRISM uses beam search with width 5 for TMM-reranked.}
\label{tab:main_results}
\centering
\small
\begin{tabular}{@{}llccccc@{}}
\toprule
 & & \multicolumn{2}{c}{Validation (in-dist, $n=10{,}000$)} & \multicolumn{3}{c}{Practical (OOD, $n=84$)} \\
\cmidrule(lr){3-4}\cmidrule(lr){5-7}
Method & Type & MAE $\downarrow$ & $R^2$ $\uparrow$ & MAE $\downarrow$ & $R^2$ $\uparrow$ & EMD $\downarrow$ \\
\midrule
SA$^\dagger$ & Optim. & 0.016 & 0.978 & \textbf{0.146} & \textbf{0.711} & 75.41 \\
Diff-TMM$^\dagger$ & Optim. & 0.031 & 0.954 & 0.204 & 0.539 & 94.86 \\
\midrule
OptoGPT & Neural & 0.059 & 0.715 & 0.356 & $-0.20$ & 97.25 \\
Tandem & Neural & 0.068 & 0.771 & 0.466 & $-0.54$ & 125.49 \\
CVAE & Neural & 0.161 & 0.066 & 0.390 & $-0.39$ & 134.67 \\
\midrule
PRISM-13M (greedy) & Neural & 0.027 & 0.945 & 0.303 & 0.247 & 96.17 \\
PRISM-13M (TMM-reranked) & Neural & 0.024 & 0.957 & 0.272 & 0.362 & 90.83 \\
\midrule
PRISM-44M (greedy) & Neural & 0.012 & 0.989 & 0.244 & 0.450 & 77.63 \\
\textbf{PRISM-44M (TMM-reranked)} & Neural & \textbf{0.010} & \textbf{0.992} & 0.228 & 0.533 & \textbf{72.20} \\
\bottomrule
\end{tabular}
\end{table*}

On the validation set, PRISM-44M greedy outperforms all methods including SA, making it the first neural method to surpass iterative optimization on this benchmark. Even the smaller PRISM-13M outperforms all neural baselines by a wide margin, despite having $4.9\times$ fewer parameters than OptoGPT (13M vs.\ 63.9M).

\subsection{Practical Targets Performance}
\label{sec:results_synthetic}
\begin{figure*}[t]
\centering
\includegraphics[width=\textwidth]{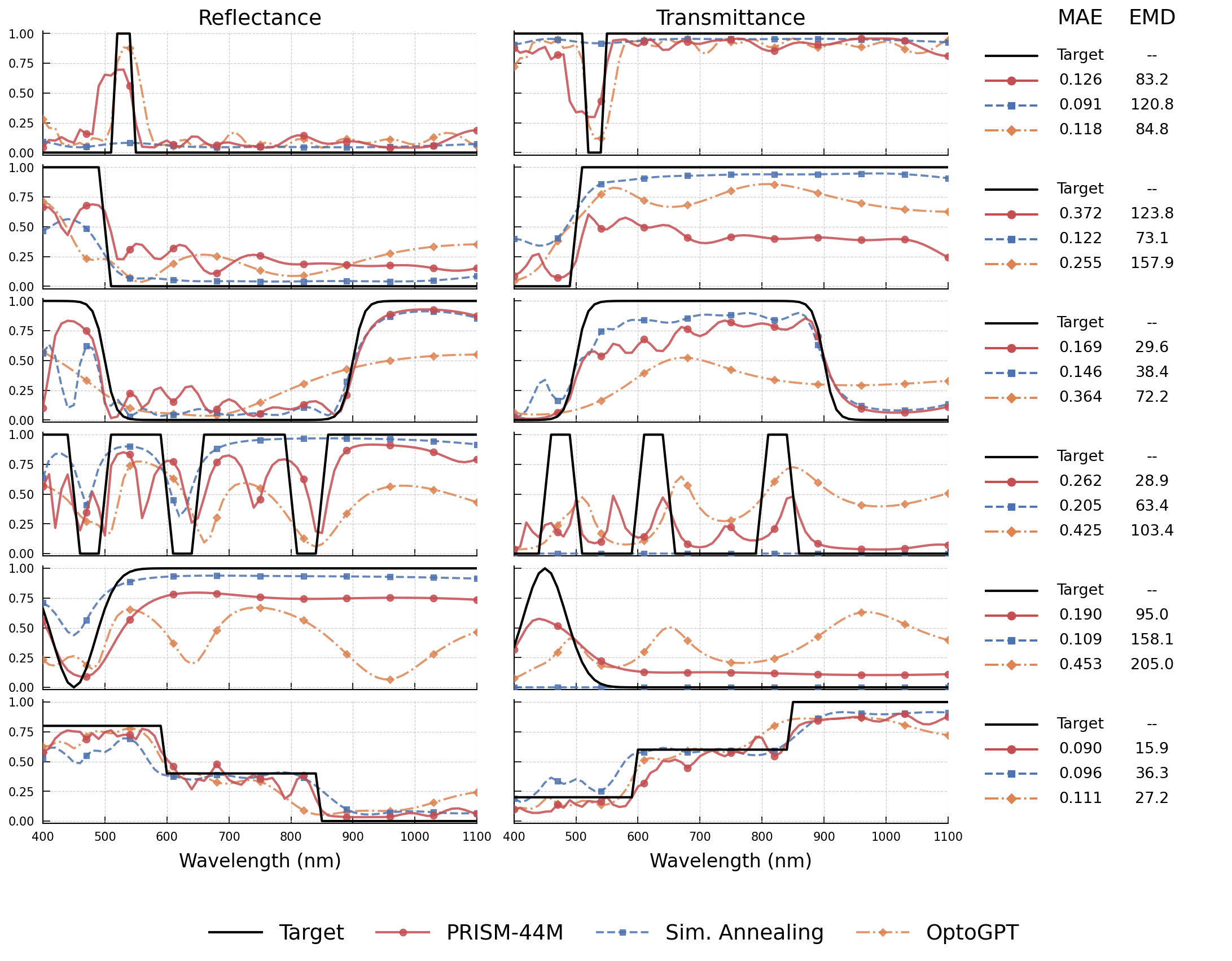}
\caption{Qualitative comparison on practical target spectra. Each panel shows a target (black) and the TMM-re-simulated spectrum of the design produced by each method. While SA achieves lower MAE on the practical benchmark overall, we can see that the solutions generated by the transformer models and PRISM follow the target curves more closely.}
\label{fig:handcrafted_comparisons}
\end{figure*}

On the practical targets, PRISM substantially outperforms all prior neural baselines on every metric. SA achieves the lowest pointwise error (MAE 0.146, $R^2$ 0.711), benefiting from direct per-target optimization, but PRISM-44M TMM-reranked achieves the lowest \emph{shape-sensitive} error: its EMD of 72.2 is below SA's 75.4 even though SA enjoys an MAE advantage. As we discuss in \Cref{sec:analysis}, pointwise MAE systematically under-credits neural methods on this benchmark. \Cref{fig:handcrafted_comparisons} shows spectral comparisons across methods on a representative subset.

\subsection{Out-of-Distribution Generalization}

\begin{figure}[!t]
\centering
\includegraphics[width=\columnwidth]{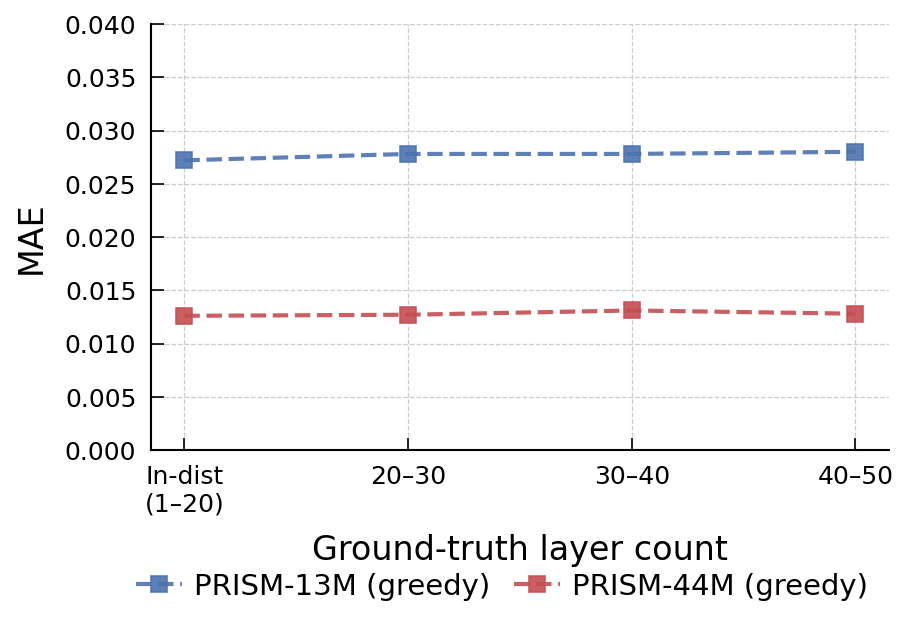}
\caption{Out-of-distribution generalization to longer sequences ($n = 10{,}000$ per condition, 10\,nm thickness steps). Both PRISM-13M and PRISM-44M exhibit minimal degradation when extrapolated to generated sequence up to $2.5\times$ longer than training.}
\label{fig:ood_generalization}
\end{figure}

\Cref{fig:ood_generalization} summarizes PRISM's performance on out-of-distribution sequence lengths for both model sizes. PRISM exhibits robust out-of-distribution generalization, maintaining greedy MAE on sequences up to 2.5× longer
than training. Scaling from 13M to 44M parameters halves
greedy MAE across all conditions, with the same qualitative behaviors at both scales.


%% file: sections/analysis.tex
\section{Analysis}
\label{sec:analysis}


On the practical benchmark, SA achieves the lowest pointwise MAE and $R^2$ (\Cref{tab:main_results}). However, qualitative inspection of the decoded spectra (\Cref{fig:handcrafted_comparisons}) tells a different story: the curves produced by the neural methods, and PRISM in particular, track the target shape more faithfully than the optimization baselines, capturing sharper band edges, peak positions, and passband widths that pointwise MAE under-credits.

The mismatch arises because MAE rewards average agreement, not structural alignment. SA's per-target optimization converges to spectra that minimize average deviation by staying close to a smoothed/mean shape, even when key structures are lost. Under EMD the ranking flips at the top. We view EMD as a critical addition to the evaluation regime for real world target cases where specific parts of the spectrum are regions of interest while the other parts are generally flat.




%% file: sections/conclusion.tex
\section{Conclusion}
\label{sec:conclusion}

We presented PRISM, an autoregressive transformer for inverse thin-film optical design that introduces two architectural innovations: spectrum prefix conditioning and cumulative-depth RoPE. A 44M-parameter model achieves greedy MAE = 0.012 on in-distribution data, surpassing all baselines including simulated annealing while running significantly faster. On 84 out-of-distribution practical targets, PRISM-44M TMM-reranked also achieves the lowest spectral earth-mover's distance, edging out simulated annealing on this shape-sensitive metric despite a higher pointwise MAE.

PRISM exhibits robust out-of-distribution generalization, maintaining greedy MAE on sequences up to $2.5\times$ longer than training. Scaling from 13M to 44M parameters halves greedy MAE across all conditions, with the same qualitative behaviours at both scales.

\paragraph{Limitations.} PRISM is conditioned on a full target spectrum across all 71 wavelengths at normal incidence. Practical filter design often involves partial specifications, such as constraining only a specific wavelength subrange while leaving the remainder free, which our current conditioning scheme does not support. Similarly, the model does not accept angle of incidence as an input, precluding designs that require angular selectivity or must perform across a range of incident angles. Both extensions would require richer conditioning interfaces and are left for future work. Beyond these, promising directions include RLVR post-training with TMM-based rewards to directly optimize spectral fidelity on practical targets, which cannot naturally occur in synthetic data.

\paragraph{Reproducibility.} All model architectures, training procedures, hyperparameters, and evaluation protocols are fully described in this paper. Code is available at \url{https://github.com/wang-henry4/prism} and pre-trained checkpoints at \url{https://huggingface.co/flying-iwik/Prism-44M-Mixed}.